\documentclass[10pt,twocolumn,letterpaper]{article}

\usepackage{iccv}
\usepackage{times}
\usepackage{epsfig}
\usepackage{graphicx}
\usepackage{subfigure}
\usepackage{amsmath}
\usepackage{amssymb}
\usepackage{algpseudocode}
\usepackage{algorithm}
\usepackage{bm}
\usepackage{enumitem}
\usepackage{multirow}

\usepackage[pagebackref=true,breaklinks=true,letterpaper=true,colorlinks,bookmarks=false]{hyperref}

\iccvfinalcopy

\setitemize[1]{itemsep=-0.2em,topsep=0em}
\usepackage[font=small]{caption}
\usepackage{setspace}
\ificcvfinal\fi
\begin{document}

    \title{\vspace{-1.5cm}Pyramid Mask Text Detector\vspace{-0.5cm}}
    \author{
    Jingchao Liu$^{1,2*}$, Xuebo Liu$^{1}$\thanks{ indicates equal contribution.}, Jie Sheng$^{1}$, Ding Liang$^{1}$, Xin Li$^{3}$, Qingjie Liu$^{2}$\\
    ${^1}$SenseTime\\
    ${^2}$Beihang University\\
    ${^3}$The Chinese University of Hong Kong\\
    \tt\small liu.siqi@buaa.edu.cn, liuxuebo@sensetime.com, jie.sheng0112@gmail.com,\\
    \tt\small liangding@sensetime.com, lixin@se.cuhk.edu.hk, qingjie.liu@buaa.edu.cn
    }

    \maketitle

    \begin{abstract}
        Scene text detection, an essential step of scene text recognition system, is to locate text instances in natural scene images automatically. Some recent attempts benefiting from Mask R-CNN formulate scene text detection task as an instance segmentation problem and achieve remarkable performance. In this paper, we present a new Mask R-CNN based framework named Pyramid Mask Text Detector (PMTD) to handle the scene text detection. Instead of binary text mask generated by the existing Mask R-CNN based methods, our PMTD performs pixel-level regression under the guidance of location-aware supervision, yielding a more informative soft text mask for each text instance. As for the generation of text boxes, PMTD reinterprets the obtained 2D soft mask into 3D space and introduces a novel plane clustering algorithm to derive the optimal text box on the basis of 3D shape. Experiments on standard datasets demonstrate that the proposed PMTD brings consistent and noticeable gain and clearly outperforms state-of-the-art methods. Specifically, it achieves an F-measure of 80.13\% on ICDAR 2017 MLT dataset.
    \end{abstract}

    \section{Introduction}\label{sec:introduction}
    \begin{figure}
  \footnotesize
  \centering
  \subfigure[Examples with imprecise segmentation labels. The area within green box denotes the manually annotated text instance. Many background pixels not belonging to the text instance are mislabeled as the foreground pixels, especially at the border of the text box, which may hurt the performance of the Mask R-CNN based methods.]{
  \begin{minipage}[c]{0.47\textwidth}
    \includegraphics[width=\textwidth, height=1.5cm]{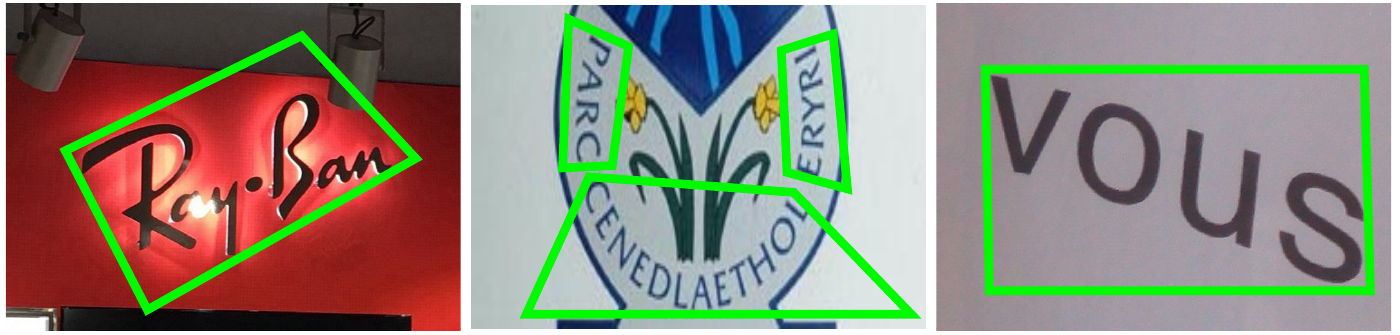}
  \label{fig:background-pixel}
  \end{minipage}}
  \subfigure[The red box is the predicted bounding box and the green box refers to the predicted text box. The existing Mask R-CNN based methods suffer from the errors of bounding box detection while PMTD can regress more accurate text box with the help of the informative soft text mask.]{
  \begin{minipage}[c]{0.47\textwidth}
    \includegraphics[width=\textwidth]{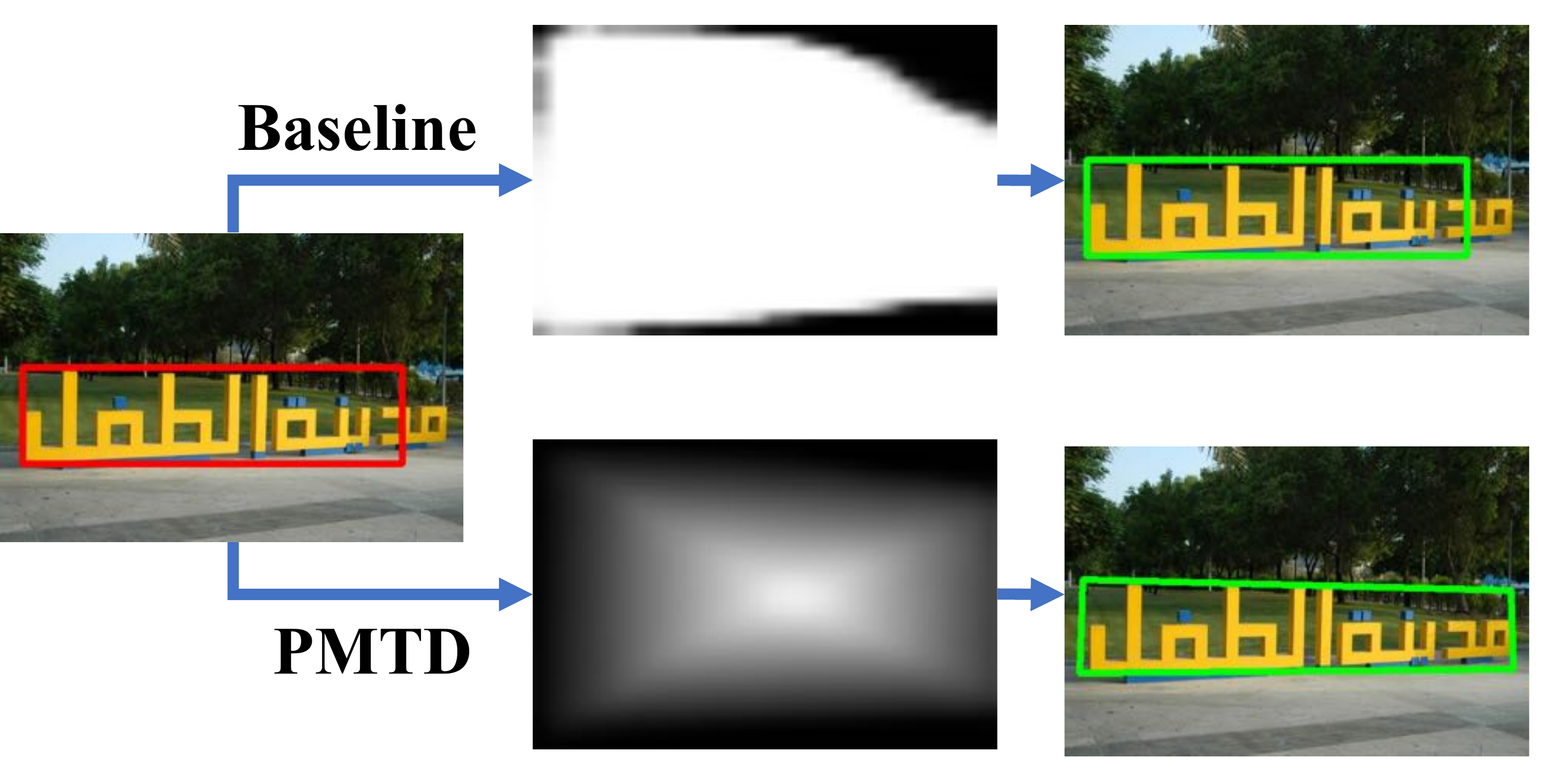}
  \label{fig:outlier}
  \end{minipage}}
  \vspace{5pt}
  \caption{Imprecise segmentation labels and imprecise bounding box are detrimental to previous methods.}
  \label{fig:drawbacks}
\end{figure}

Scene text detection has attracted growing research interests in the computer vision community, due to its numerous practical applications in scene understanding, license plate recognition, autonomous driving, and document analysis, etc. Recently, many works~\cite{xie2018scene, lyu2018mask, huang2018mask} view scene text detection as an instance segmentation problem, and several Mask R-CNN~\cite{he2017mask} based methods were proposed and achieved remarkable performance. However, there are several drawbacks in these works:

\textbf{Over-simplified supervision}: The common observation that most text areas in natural scenes are quadrilateral is supposed to be useful for text detection. However, the Mask R-CNN based methods, aiming for differentiating the text region from the background region rather than generating a text mask of a specific shape, ignore the consideration of such kind of information and therefore can not take advantage of the given label.

\textbf{Imprecise segmentation labels}: Converting the quadrilateral text area into a pixel-level binary supervision signals for semantic segmentation enables directly applying Mask R-CNN to scene text detection. However, the quality of the generated pixel-level labels is unsatisfactory. As shown in Fig.\ref{fig:background-pixel}, many background pixels not belonging to the text region are incorrectly regarded as the foreground pixels. Trained on noisy data, the semantic segmentation based text detector is prone to generate mistakes.

\textbf{Error propagation}: The Mask R-CNN based methods firstly predict text bounding boxes and then perform semantic segmentation within the bounding box. Such strategy is usually reasonable for simple scenes but rather fragile when the predicted bounding box fails to cover the whole text region. The reason is because determining the text box with only the text region inside the bounding box tends to exclude the outside part (See Fig.~\ref{fig:outlier}). In other words, the errors from the object detection may be propagated to the process of finding text box, leading to performance degradation of scene text detection. We also observe that the effect of the error propagation will be amplified with the increasing of the IoU threshold for true positive text instances (quantitative results and qualitative analysis are detailed in Sec.~\ref{subsec:ablation-study}).

\begin{figure}
    \includegraphics[width=0.47\textwidth]{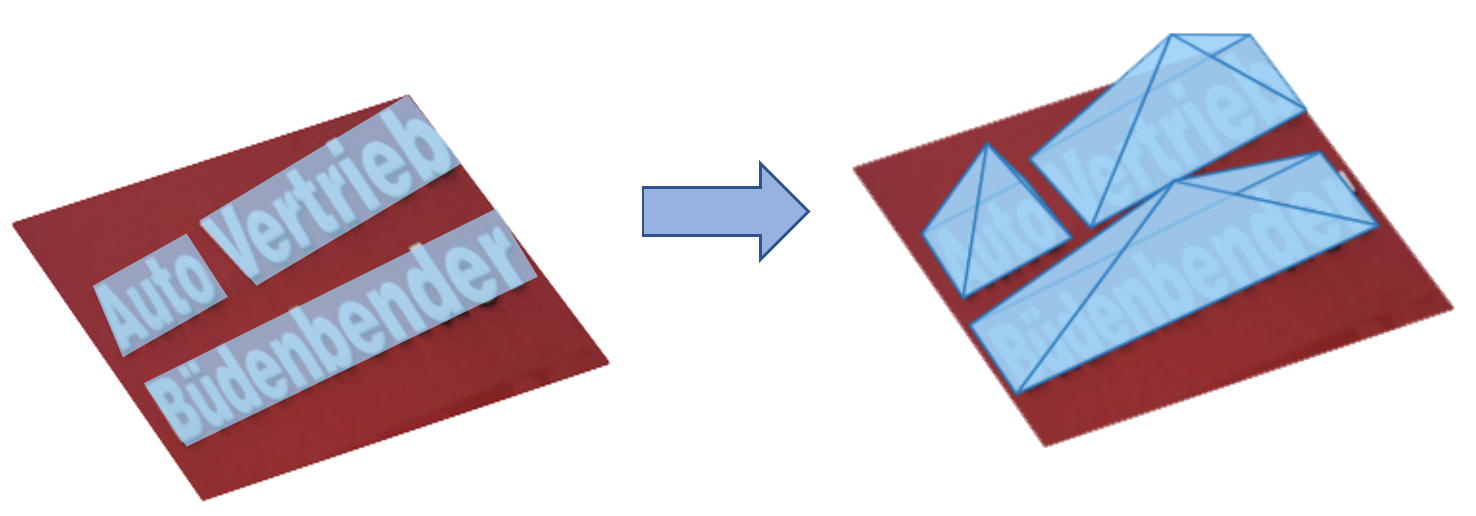}
    \vspace{5pt}
    \caption{Previous methods aim to find \{0, 1\} label for each pixel while PMTD assigns a soft pyramid label of the value $\in$ [0, 1].}
    \label{fig:pyramid-mask}
\end{figure}

In this paper, we propose the Pyramid Mask Text Detector (PMTD) to address the above problems. As depicted in Fig.~\ref{fig:pyramid-mask}, instead of pixel-level binary classification as done in the existing Mask R-CNN based methods, we propose to perform ``soft'' semantic segmentation between the text region and the background region. Explicitly, we assign a soft pyramid label (i.e., a real value between 0 and 1) for each pixel within text instance. The value of the soft pyramid label is determined by the distance to the boundary of the text box, which implicitly encoding the shape and location information into the training data. By fitting such soft text mask, the quadrilateral property of the text instance is naturally considered during training. Besides, introducing the location-aware segmentation labels reduces the impact of mislabeled pixels near the boundary of the text box.

During the test phase, with the extended $z$-axis characterizing the value of the pixel-level segmentation output, we reinterpret the 2D predicted text mask into a set of 3D points. A plane clustering algorithm is proposed to regress the optimal pyramid from these 3D points. Specifically, launched with four initialized supporting planes of a pyramid, the plane clustering algorithm iteratively groups the nearest points for each supporting planes, and then updates the supporting planes by the clustered points. After the iterations, an accurate bounding pyramid is obtained and its bottom face is regarded as the output text box. Since it is not the boundary pixels but the supporting plane that gets involved in finding the text box, the error propagation issue can be alleviated, and more accurate text box can be obtained.

\begin{figure*}
    \includegraphics[width=\textwidth]{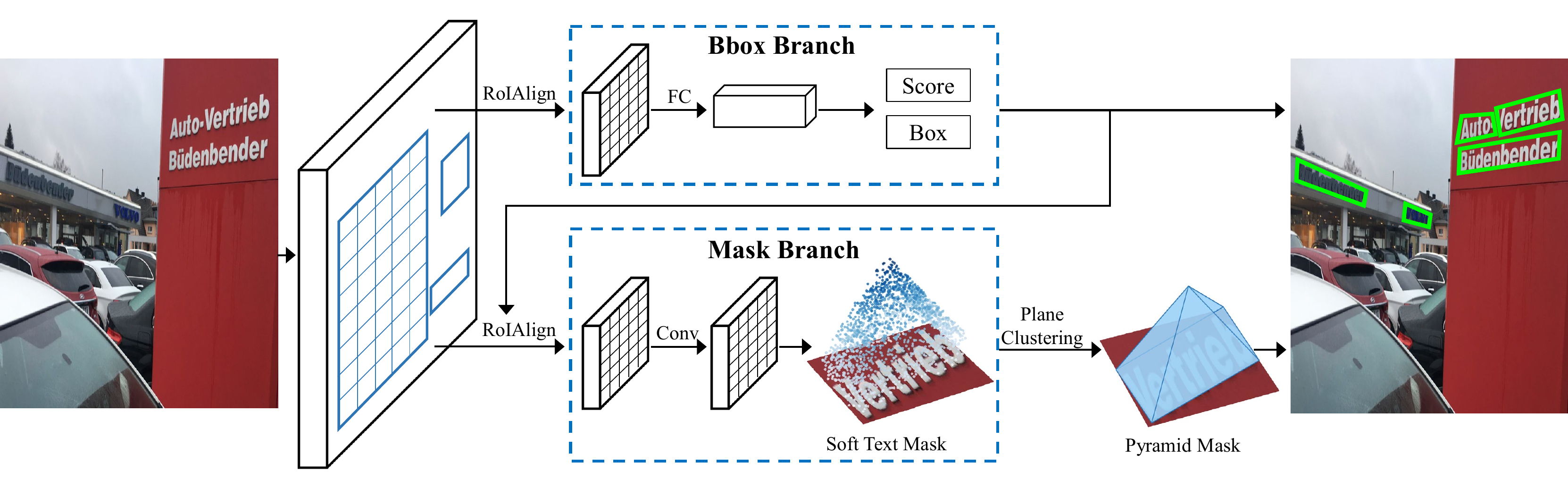}
    \vspace{0pt}
    \caption{Overall architecture of PMTD.}
    \label{fig:architecture}
\end{figure*}

Our pipeline is shown in Fig.\ref{fig:architecture}. As there exist differences between text detection datasets and object detection datasets, such as the different distribution of aspect ratios and scales, we tailor-make a Mask R-CNN based baseline for text detection, which outperforms all previous methods on ICDAR 2017 MLT dataset. Furthermore, the proposed PMTD raises the F-measure to 80.13\%. The main contributions of this paper are three-fold:

\begin{itemize}
    \item We propose the Pyramid Mask Text Detector for scene text detection, and extensive experiments demonstrate its state-of-the-art performance on several benchamark datasets.
    \item We propose to perform ``soft'' segmentation between text region and non-text region, incorporating the shape and location information into the model training and alleviating the inaccuracy labeling for the instance boundary.
    \item We introduce a novel plane clustering algorithm to find better text box with the 3D coordinate, which predicts more accurate text box and improves the robustness to imprecise bounding box predictions.
\end{itemize}

    \section{Related Work}\label{sec:related-work}
    Scene text detection has received significant attention over the past few years, and numerous deep learning based methods~\cite{gupta2016synthetic,zhang2018feature,liao2018textboxes++,ma2018arbitrary,zhou2017east,he2017deep,li2018shape,long2018textsnake,xie2018scene} have been reported in the literature. Comprehensive reviews and detailed analyses can be found in survey papers~\cite{zhu2016scene,uchida2014text,ye2015text}. 

Earlier text detection works including~\cite{huang2014robust,jaderberg2014deep,jaderberg2016reading} are among the first deep neural network based methods.
They usually consist of multiple stages, such as candidate aggregation, word partition and false positive removal by post-processing filtering. Huang \etal~\cite{huang2014robust} first apply the MSERs operator on the input image to generate some text candidates, then use a CNN classifier to generate a confidence map which was later used for constructing text-lines. Jaderberg \etal~\cite{jaderberg2014deep} train a strongly supervised character classifier to generate text saliency map, then combines bounding boxes at multiple scales and undergoes filtering and non-maximal suppression. In a later work~\cite{jaderberg2016reading}, they leverage a CNN for bounding box regression and a random forest classifier for reducing the number of false-positive detections.

Recent works~\cite{gupta2016synthetic,tian2016detecting,liao2017textboxes,zhang2018feature} regard text words or lines as objects and adapt the pipeline of general object detection, e.g., Faster R-CNN~\cite{ren2015faster}, SSD~\cite{liu2016ssd} and YOLO~\cite{redmon2016you} into text detection. They regress the offsets from a proposal region or a single pixel in the feature map to a horizontal rectangle and obtain good performance with well-designed modifications on horizontal text detection. Gupta \etal~\cite{gupta2016synthetic} improves over the YOLO network and Fully Convolutional Networks (FCN)~\cite{long2015fully} for text prediction densely, while further adopts the filter and regression steps for removing the false positives. TextBoxes~\cite{liao2017textboxes} modifies SSD by using irregular convolutional kernels and long default anchors according to the characteristic of scene text. Built on top of Faster R-CNN, CTPN~\cite{tian2016detecting} develops a vertical anchor mechanism that predicts location and text/non-text score of each fixed-width proposal simultaneously, then connects the sequential proposals by a recurrent neural network. FEN~\cite{zhang2018feature} improves text recall with a feature enhancement RPN and hyper feature generation for text detection refinement.

Considering that scene texts are with arbitrary orientations, works in~\cite{liao2018textboxes++,ma2018arbitrary,zhou2017east,he2017deep,liu2018fots,he2018end} make the above methods possible for multi-oriented text detection. RRPN~\cite{ma2018arbitrary} introduces inclined anchors with angle information for arbitrary-oriented text prediction and rotated RoI pooling layer to project arbitrary-oriented proposals to the feature map for a text region classifier. TextBoxes++~\cite{liao2018textboxes++} improves TextBoxes by regressing horizontal anchors to more general quadrilaterals enclosing oriented texts. It also proposes an efficient cascaded non-maximum suppression for quadrilaterals or rotated rectangles. With dense predictions and one step post processing, EAST~\cite{zhou2017east} and DDR~\cite{he2017deep} both directly produce the rotated boxes or quadrangles of text at each point in the text region. Recent text spotting methods like FOTS~\cite{liu2018fots} and He \etal~\cite{he2018end} show that training text detection and recognition simultaneously could greatly boost detection performance.

Except for the above regression-based methods,~\cite{deng2018pixellink,li2018shape,long2018textsnake,xie2018scene} cast text detection as a segmentation problem. PixelLink~\cite{deng2018pixellink} first segments out text regions by linking pixels within the same instance, then extracts text bounding boxes directly from the segmentation without location regression.
TextSnake~\cite{long2018textsnake} employs an FCN~\cite{long2015fully} model to estimate the geometry attributes of text instances and uses a striding algorithm to extract the central axis point lists and finally reconstruct the text instances.
Segmentation based methods tend to link adjacent text regions together incorrectly. To address this problem, PSENet~\cite{li2018shape} finds text kernels with various scales and proposes a progressive scale expansion algorithm to separate text instances standing close to each other accurately.
SPCNET~\cite{xie2018scene} views text detection as an instance segmentation problem, based on Mask R-CNN, it proposes a text context module and a re-score mechanism to suppress false positives.

Although Curved text detection~\cite{ch2017total, yuliang2017detecting} has attracted growing research interests recently, quadrilateral text detection is still a fundamental and challenging problem to be solved. PMTD is designed specially for quadrilateral text detection and significantly improves the state-of-the-art result from 74.3\%~\cite{huang2018mask} to 80.13\% on ICDAR 2017 MLT.

    \section{Methodology}\label{sec:methodology}
    In this section, we firstly introduce a strong baseline. Then the soft pyramid label is proposed, which encodes the shape and location information into the training data. Finally, a new boundary regression algorithm, namely, plane clustering, is introduced to find the most fitting pyramid of the predicted soft text mask.

\subsection{Our Baseline}\label{subsec:our-baseline}
Our baseline is based on Mask R-CNN with ResNet50 backbone~\cite{he2016deep}. In the training stage, we treat the axis-aligned bounding rectangle of the text region as the ground-truth bounding box and assign pixels inside text boundary to positive segmentation label. In the test stage, we firstly find all the connected areas in the predicted mask, then select the one with the maximum area, and finally obtain the output text box by finding the minimum bounding rectangle of this connected area.

We design a strong baseline by making the following three modifications:

\textbf{Data augmentation}: To enhance the generalization ability to various scales and aspect ratios, we apply data augmentations to enlarge scene text datasets:
\begin{enumerate}[nosep]
    \item Random horizon flip with a probability of 0.5.
    \item Random resize the height and width of images to 640-2560 individually, without keeping the original aspect ratio.
    \item Random select one $640\times640$ crop region from the resized image.
\end{enumerate}

\textbf{RPN Anchor}: When adopting the FPN module, we can quantify the anchor by three parameters: the base scale of anchors, the feature maps where anchors searched, and the aspect ratios of anchors.

First of all, based on statistics of the data-augmented ground truth bounding box's height and width, we set the base scale of the anchor to $4\times4$ among all the four feature maps $\{1/4, 1/8, 1/16, 1/32\}$ uniformly.

For the anchor's aspect ratio, we calculate out five dedicated aspect ratios: $\{0.17, 0.44, 1.13, 2.90, 7.46\}$. The detail of generating aspect ratios is as follows: first, analyze the data-augmented ground truth bounding box's aspect ratio, then get the 5\% quantile 0.17 and 95\% quantile 7.46, finally insert three values in equal proportion between the 5\% and 95\% quantiles to form the final aspect ratio list.

\textbf{OHEM}: In the bounding box branch, we adopt the OHEM~\cite{shrivastava2016training} to learn the hard samples. In our settings, we first sort the samples provided by RPN in the descending order of the sum of classification loss and location loss, then select the top 512 difficult samples to update the network.

\subsection{Motivation}\label{subsec:motivation}

Although our baseline achieves remarkable performance, it still has the same drawbacks as other Mask R-CNN based methods, as mentioned in Sec.~\ref{sec:introduction}:

\begin{itemize}
    \item These methods are not considering the common observation that most text areas in natural scenes are quadrilateral. They break down the quadrilateral structure into a pixle-wise classification problem which losses the shape information of the mask.
    \item Converting the quadrilateral text areas into pixel-level supervision is imprecise. Many background pixels not belonging to the text region are incorrectly regarded as the foreground pixels, as shown in Fig.~\ref{fig:background-pixel}. The mislabeled boundary pixels may cause an unexpectedly misjudged loss.
    \item Mask R-CNN based methods firstly predict bounding boxes and then predict text mask for every bounding box. The imprecisely predicted bounding box limits the mask branch to generate accurate text mask. In other words, the errors from the object detection will be propagated to the following steps, as shown in Fig.~\ref{fig:outlier}.
\end{itemize}

These problems motivate us to build pyramid mask text detector (PMTD), a new pipeline for scene text detection. The PMTD predicts a soft text mask for each text region and apply plane clustering algorithm to convert the predicted soft mask to the pyramid mask.

\begin{figure}
    \includegraphics[width=0.47\textwidth]{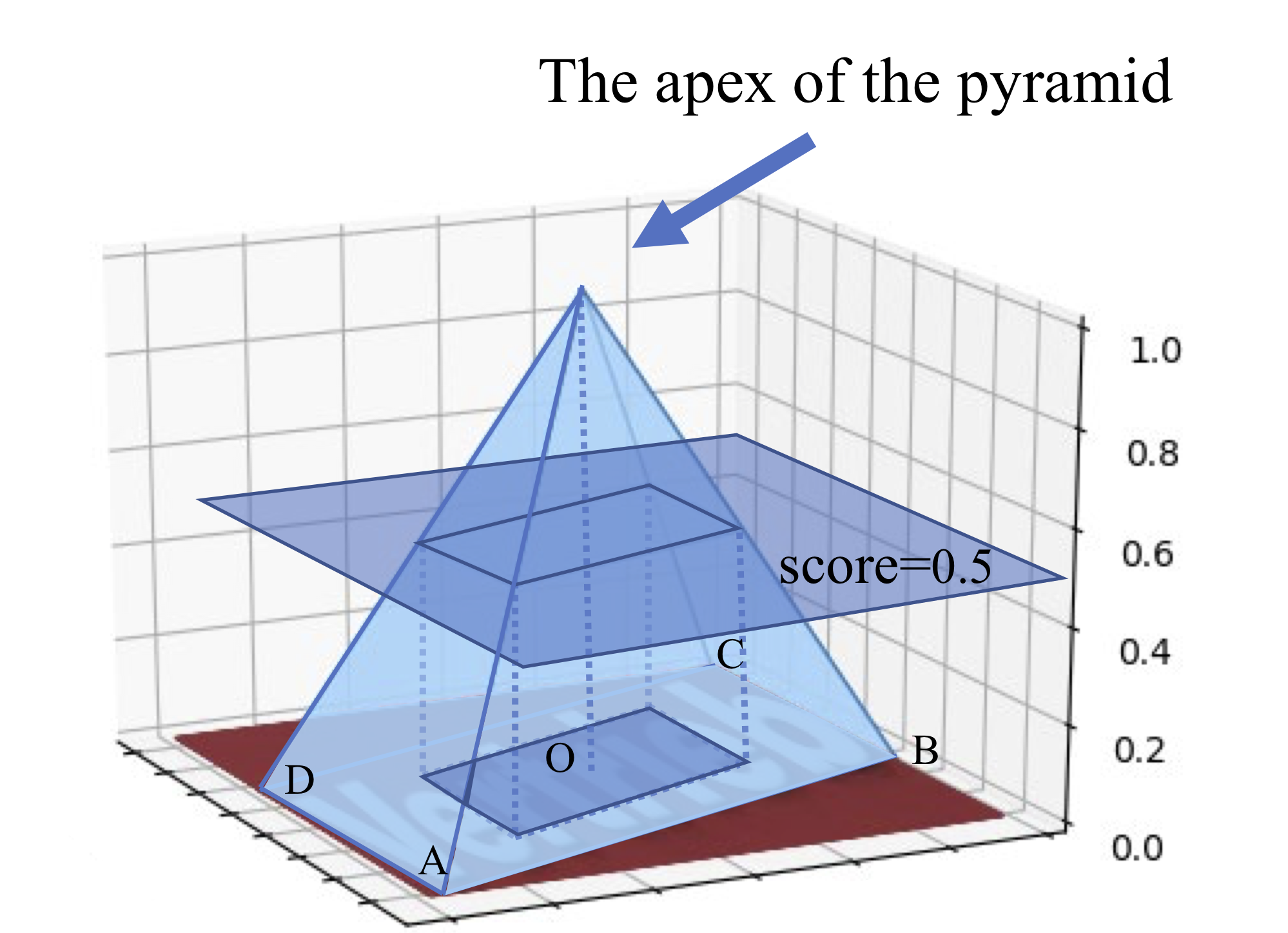}
    \vspace{5pt}
    \caption{Generation of soft pyramid label. For a pixel in the text area, its label is the height of the pyramid.}
    \label{fig:pm_generate}
\end{figure}

\subsection{Pyramid Label}\label{subsec:pyramid-mask}
We refines the mask's hard label of the class $\in$ $\{0, 1\}$ to the soft pyramid label of the score $\in$ $[0, 1]$ so that the PMTD can capture the shape and location information from the data. Specifically, we assign the center of text region as the apex of the pyramid with an ideal value $score=1$ and the boundary of text region as the bottom edge of the pyramid. We use the linear interpolation to fill each triangle side of the pyramid, as illustrated in Fig.\ref{fig:pm_generate}.

Formally, given the four corner points $A(x_{a}, y_{a})$, $B(x_{b}, y_{b})$, $C(x_{c}, y_{c})$, $D(x_{d}, y_{d})$ of a quadrilateral, the value $score_{p}$ for the point $P(x_{p}, y_{p})$ can be calculated as follows. First, the center of text region $O(x_{o}, y_{o})$ can be obtained by:
\begin{align}
    x_{o}&= (x_{a}+x_{b}+x_{c}+x_{d}) / 4 \\
    y_{o}&= (y_{a}+y_{b}+y_{c}+y_{d}) / 4
\end{align}
For every region $R_{OMN}$ (region between two rays $OM$ and $ON$) from $R_{OAB}$, $R_{OBC}$, $R_{OCD}$, $R_{ODA}$, the $\overrightarrow{OP}$ can be decomposed uniquely:
\begin{align}
    \overrightarrow{OP} &= \alpha\overrightarrow{OM}+\beta\overrightarrow{ON} \\
    \begin{bmatrix}x_{p}-x_{o} \\ y_{p}-y_{o} \end{bmatrix} &= \begin{bmatrix} x_{m}-x_{o} & x_{n}-x_{o} \\ y_{m}-y_{o} & y_{n}-y_{o} \end{bmatrix} \begin{bmatrix}\alpha \\ \beta \end{bmatrix}
    \end{align}
    Then, $\alpha$ and $\beta$ can be obtained by
    \begin{align}
    \begin{bmatrix}\alpha \\ \beta \end{bmatrix} &= \begin{bmatrix} x_{m}-x_{o} & x_{n}-x_{o} \\ y_{m}-y_{o} & y_{n}-y_{o} \end{bmatrix}^{-1} \begin{bmatrix}x_{p}-x_{o} \\ y_{p}-y_{o} \end{bmatrix}
\end{align}
The region $R$ which $P$ belongs to needs to satisfy the following condition:
\begin{equation}
    \alpha \ge 0\ and\ \beta \ge 0
\end{equation}
Then the $score_{p}$ can be calculated by:
\begin{equation}
    score_{p} = \bm{\max}(1 - (\alpha+\beta), 0)
\end{equation}

During the training stage, such supervision is reasonable. If one pixel locates near the center of the instance, its receptive field will be filled with positive pixels and deserves a higher score consequently. While the receptive field of the pixels near the boundary will contain much background context, and the scores of these pixels should be close to 0. In this respect, a larger receptive field is vital for PMTD to attain more precise results. So in the mask head, we replace the first four convolution layers to dilated convolution with stride 2 to enlarge receptive field.

\begin{figure}
    \includegraphics[width=0.47\textwidth]{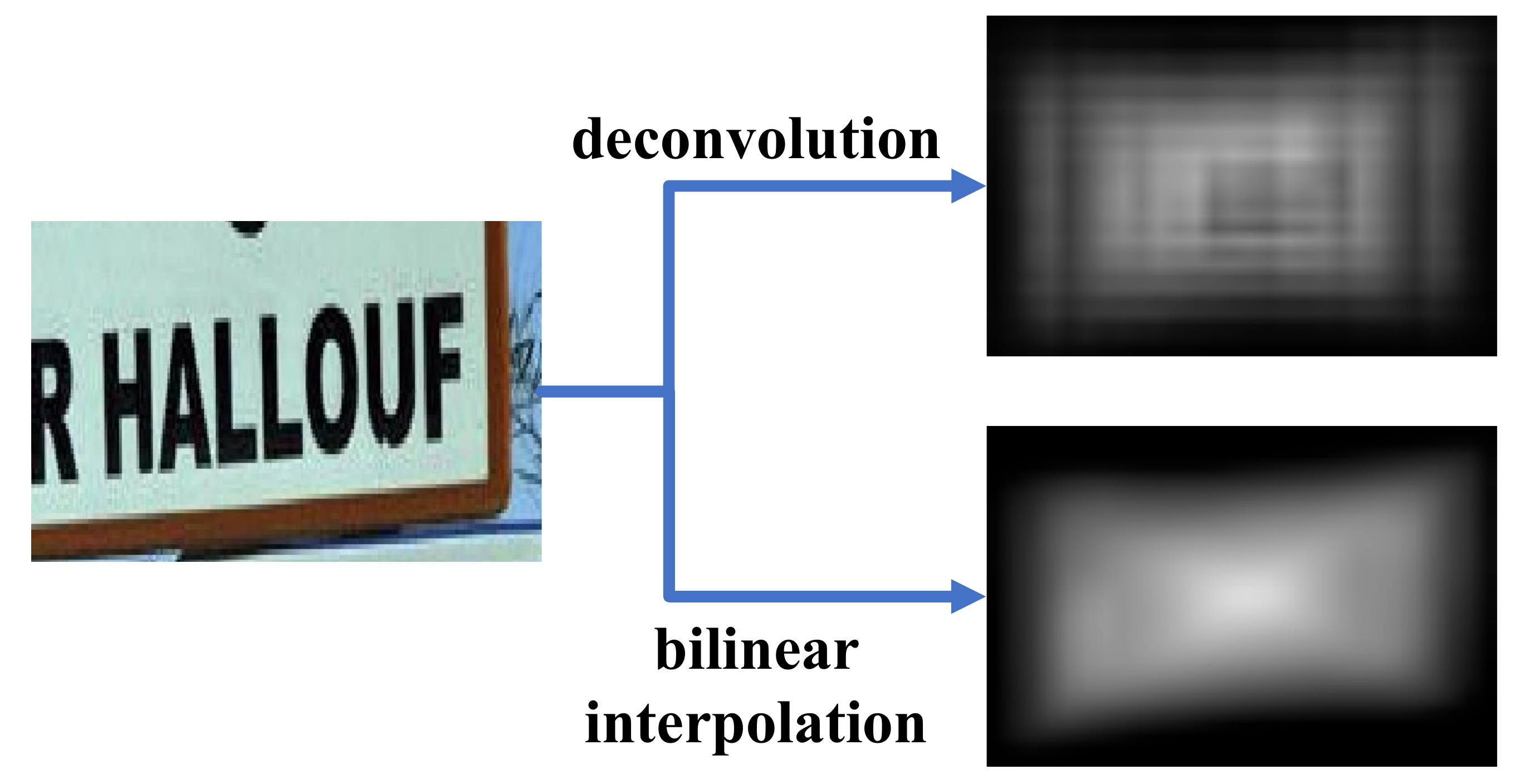}
    \vspace{5pt}
    \caption{Deconvolution causes checkerboard pattern in our experiment, so we use bilinear interpolation for upsample to get a more accurate mask.}
    \label{fig:transposed-conv}
\end{figure}

Moreover, as mentioned in~\cite{odena2016deconvolution}, deconvolution may cause the checkerboard pattern, which is harmful to the pixelwise regression, as illustrated in Fig.~\ref{fig:transposed-conv}. To avoid this, we replace the deconvolution layer in the mask head to bilinear interpolation and a followed convolution layer.

We employ pixelwise $L_{1}$ loss to optimize the predicted text mask. Following the design in Mask R-CNN, the loss function of the whole network is as follows:
\begin{equation}
    L = L_{\text{rpn}} + \lambda_{\text{1}} L_{\text{cls}} + \lambda_{\text{2}} L_{\text{box}} + \lambda_{\text{3}} L_{\text{pyramid\_mask}}
\end{equation}
$\lambda_{\text{1}}$, $\lambda_{\text{2}}$ and $\lambda_{\text{3}}$ are set to 1, 1, 5 respectively in our experiments.

Training with this new style label alleviates the pixel mislabeling problem. Taking a background pixel near the boundary as an example. Although it is mistakenly regarded as the foreground, its ground truth in our methods is still close to 0, while in previous Mask R-CNN based methods, this pixel is labeled as 1.

\subsection{Plane Clustering}\label{subsec:Plane-clustering}
In this section, we will illustrate the plane clustering algorithm in details, which is an iteratively updated clustering algorithm for regressing the most fitting text box from the predicted soft text mask.

As a reverse process of generating a pyramid label from text region, we will first construct the pyramid from the text mask, then take the bottom edge of the pyramid as the output text box. Hence, the critical point is to parameterize and rebuild the pyramid.

Formally, the pyramid is composed of four supporting planes and one base plane. In the context of pyramid mask, we can convert the predicted soft mask into a point set of $(x, y, z)$, in which the $(x, y)$ denotes the location and $z$ stands for the predicted score of this pixel. The base plane is formulated as the plane $z=0$, and each supporting plane can be uniquely determined by the equation  $Ax+By+Cz+D=0, C = 1$. Consequently, the task of the plane clustering algorithm is reduced to find the optimal parameter $\{A, B, D\}$ for each supporting plane, see Algorithm~\ref{alg1:plane-clustering} for details.

\begin{figure}
    \includegraphics[width=0.47\textwidth, height=2cm]{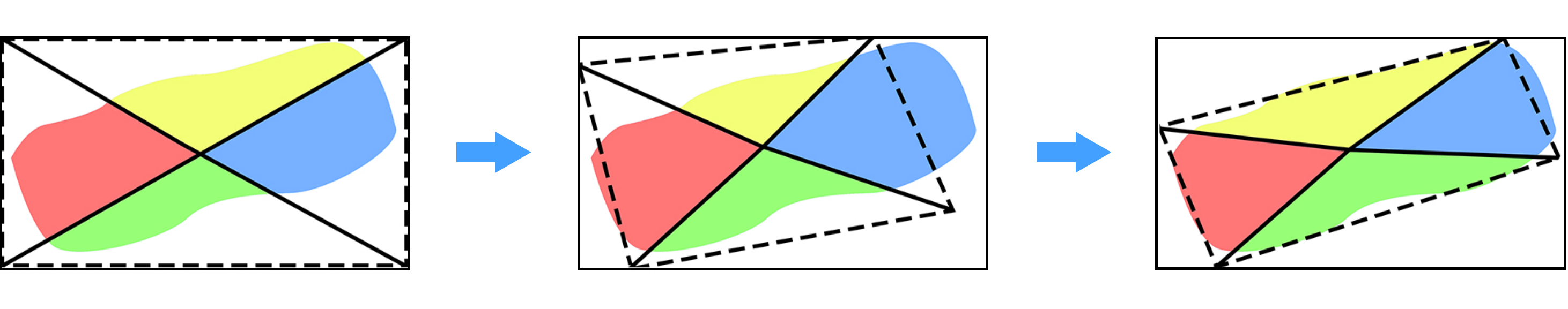}
    \vspace{5pt}
    \caption{Illustration of the plane clustering algorithm. Every positive point in the predicted soft mask is assigned to one of four different colors, which indicates the supporting plane the point belongs to. The dashed lines are the intersection of supporting planes and the bottom plane, which form the predicted text box together. The supporting planes are refined from left to right.}
    \label{fig:pc}
\end{figure}

\renewcommand{\algorithmicrequire}{~~~~\textbf{input:}}
\renewcommand{\algorithmicensure}{~~~~\textbf{output:}}
\begin{algorithm}
    \caption{Plane Clustering}
    \label{alg1:plane-clustering}
    \begin{algorithmic}
        \Require \\
        ~~$point:~location = (x, y), predicted\_score = z$ \\
        ~~$points = [point\_num = H * W, (x, y, z)]$
        \Ensure \\
        ~~$plane:~Ax + By + Cz + D = 0, C = 1$ \\
        ~~$planes = [plane\_num = 4, (A, B, D)]$ \\
        \Function{Plane Clustering}{$points$}
        \State $P \gets \textproc{SelectPositive}(points)$
        \State $apex.(x, y) \gets \textproc{mean}(P).(x, y)$
        \State $apex.z \gets 1$
        \State $planes \gets \textproc{InitPlanes}(apex)$
        \While{$iter<max\_iter\ and\ \textproc{Reject}(residuals)$}
        \State $G \gets \varnothing \times plane\_num$
        \For{$p \in P$}
        \State $plane\_i \gets \textproc{NearestPlane}(p, planes)$
        \State $G[plane\_i] \gets G[plane\_i] \cup \{p\}$
        \EndFor
        \State $planes, residuals \gets \textproc{RLS}(G)$
        \EndWhile
        \State \textbf{return} planes
        \EndFunction
    \end{algorithmic}
\end{algorithm}

In the initialization stage, the positive points set $P$ is built by the condition $z>0.1$. Then the apex of the initial pyramid is assigned as the center of $P$, with an ideal score $z=1$. The four vertexes of the pyramid in the bottom face are initialized as four corner points of the predicted text bounding box, shown in the left image in Fig.\ref{fig:pc}.

After initializing the pyramid, an iterative updating scheme is implemented for clustering points, which is shown in Fig.\ref{fig:pc}. In the assignment step, we partition each point to the nearest plane, and in the update step, we employ the robust least square algorithm (RLS)~\cite{holland1977robust} to regress four supporting planes from the clustered points respectively, which is robust to the noise in the predicted text mask.

When the iteration reaches the max iteration or the regression residuals returned by RLS is small enough, the final quadrangular pyramid is obtained. Then the text box can be calculated out by the intersection of four supporting planes and the plane $z=0$. In our experiment, the max iteration and the residual threshold is assigned to 10 and 1e-4 respectively.

Thanks to the more informative soft text mask, the plane clustering algorithm takes advantage of the whole soft mask's information to regress the most fitting pyramid. As the final text box is obtained from the supporting plane rather than the boundary pixels, PMTD is robust to imprecise bounding boxes, and naturally regress more accurate text boundary, detailed in Sec.~\ref{subsec:ablation-study}.

    \section{Experiments}\label{sec:experiments}
    In this section, We evaluate our approach on ICDAR 2017 MLT~\cite{nayef2017icdar2017}, ICDAR 2015~\cite{karatzas2015icdar15} and ICDAR 2013~\cite{karatzas2013icdar}. Experiment results demonstrate that the proposed PMTD brings consistent and noticeable gain, and clearly outperforms the state-of-the-art methods. Furthermore, the ablation study shows PMTD is robust to imprecise bounding box predictions and predicts more accurate text boxes.

\begin{table}
  \small
  \begin{center}
  \begin{tabular}{l|ccc}
  \hline
  Method    & Precision   & Recall   & F-measure \\ \hline
  FOTS~\cite{liu2018fots} & 80.95&57.51&67.25   \\
  FOTS$^*$~\cite{liu2018fots}& 81.86&62.30&70.75    \\
  Lyu \etal~\cite{lyu2018multi}& 83.80&55.60 &66.80    \\
  Lyu \etal$^*$~\cite{lyu2018multi}& 74.30&70.60&72.40    \\
  PSENet~\cite{li2018shape}& 77.01&68.40&72.45    \\
  Pixel-Anchor~\cite{li2018pixel} & 79.54&59.54&68.10 \\
  Pixel-Anchor$^*$~\cite{li2018pixel} & 83.90&65.80&73.76 \\
  SPCNET~\cite{xie2018scene}&  66.90&73.40&70.00    \\ 
  SPCNET$^*$~\cite{xie2018scene}& 68.60&80.60&74.10    \\ 
  Huang \etal~\cite{huang2018mask} &80.00 &69.80& 74.30 \\ \hline
  Baseline & 84.72 &	70.37 &	76.88  \\
  PMTD & 85.15&72.77&	78.48   \\ 
  PMTD$^*$ & 84.42&	76.25&	\textbf{80.13}  \\ \hline
  \end{tabular}
  \end{center}
  \caption{Comparison with other results on ICDAR 2017 MLT. $^*$ means multi scale testing.}
  \label{tab:ic17}
\end{table}

\subsection{Datasets}\label{subsec:datasets}

\textbf{ICDAR 2017 MLT} is a multi-oriented, multi-scripting, and multi-lingual scene text dataset. It consists of 7200 training images, 1800 validation images, and 9000 test images. The text regions are annotated by four vertices of the quadrilateral. It is one of the largest and most challenging scene text detection datasets.

\textbf{ICDAR 2015} is another multi-oriented text detection dataset only for English, which includes 1000 training images and 500 testing images. Similar to ICDAR 2017 MLT, the text region is also annotated as a quadrilateral.

\textbf{ICDAR 2013} is a dataset that points at the horizontal text in the natural scene. This dataset consists of 229 training images and 233 testing images.

\subsection{Comparisons with Other Methods}\label{subsec:comparisons-with-other-methods}

In this section, we compare PMTD with state-of-the-art methods on standard datasets. As shown in Tab.~\ref{tab:ic17},~\ref{tab:ic15},~\ref{tab:ic13}, our method outperforms others in all datasets.

\begin{figure*}
    \centering
    \subfigure[ICDAR 2017 MLT]{
    \begin{minipage}[c]{0.3\textwidth}
      \includegraphics[width=\textwidth, height=4cm]{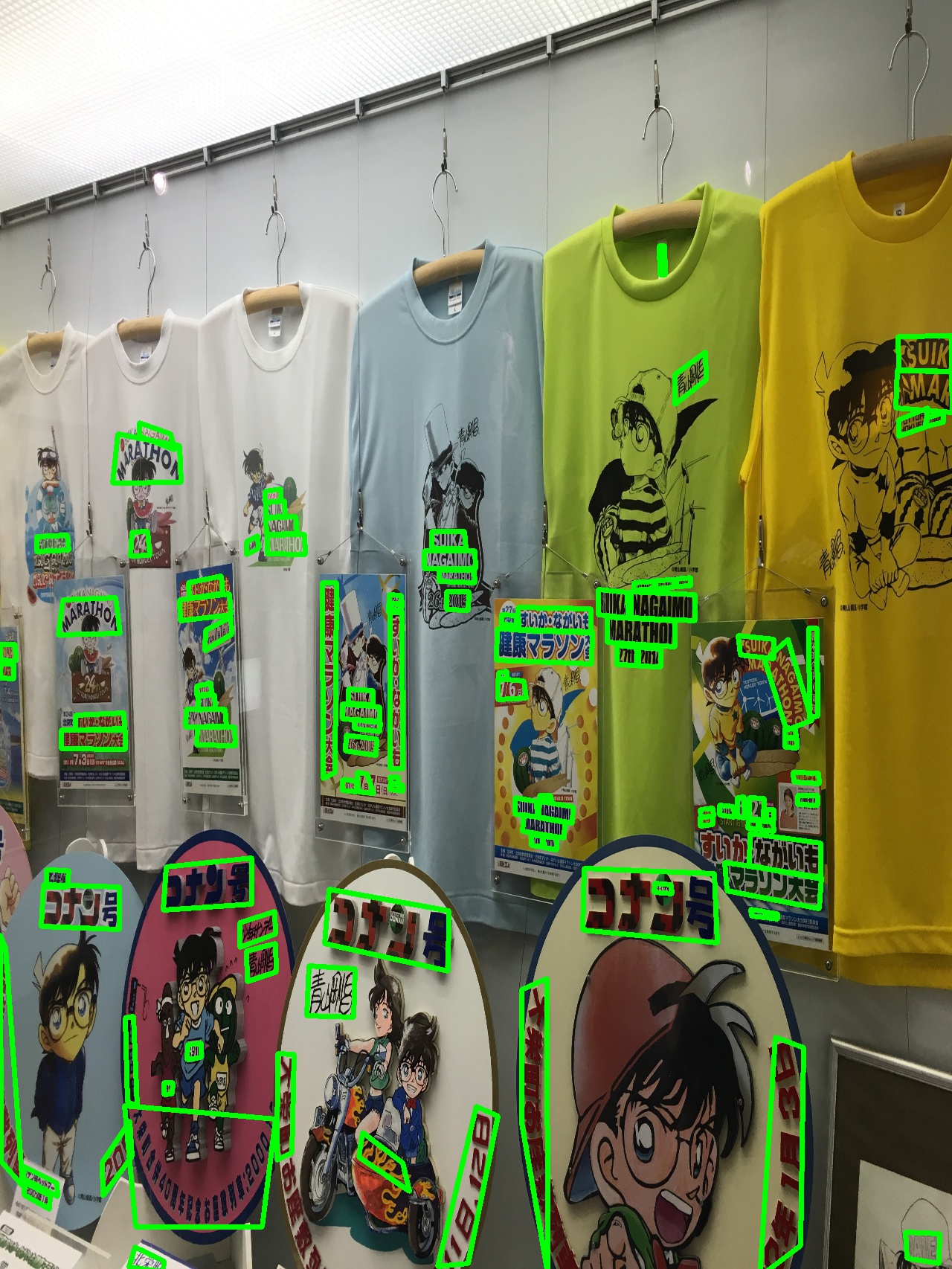}
      \vspace{5pt}
      \includegraphics[width=\textwidth, height=4cm]{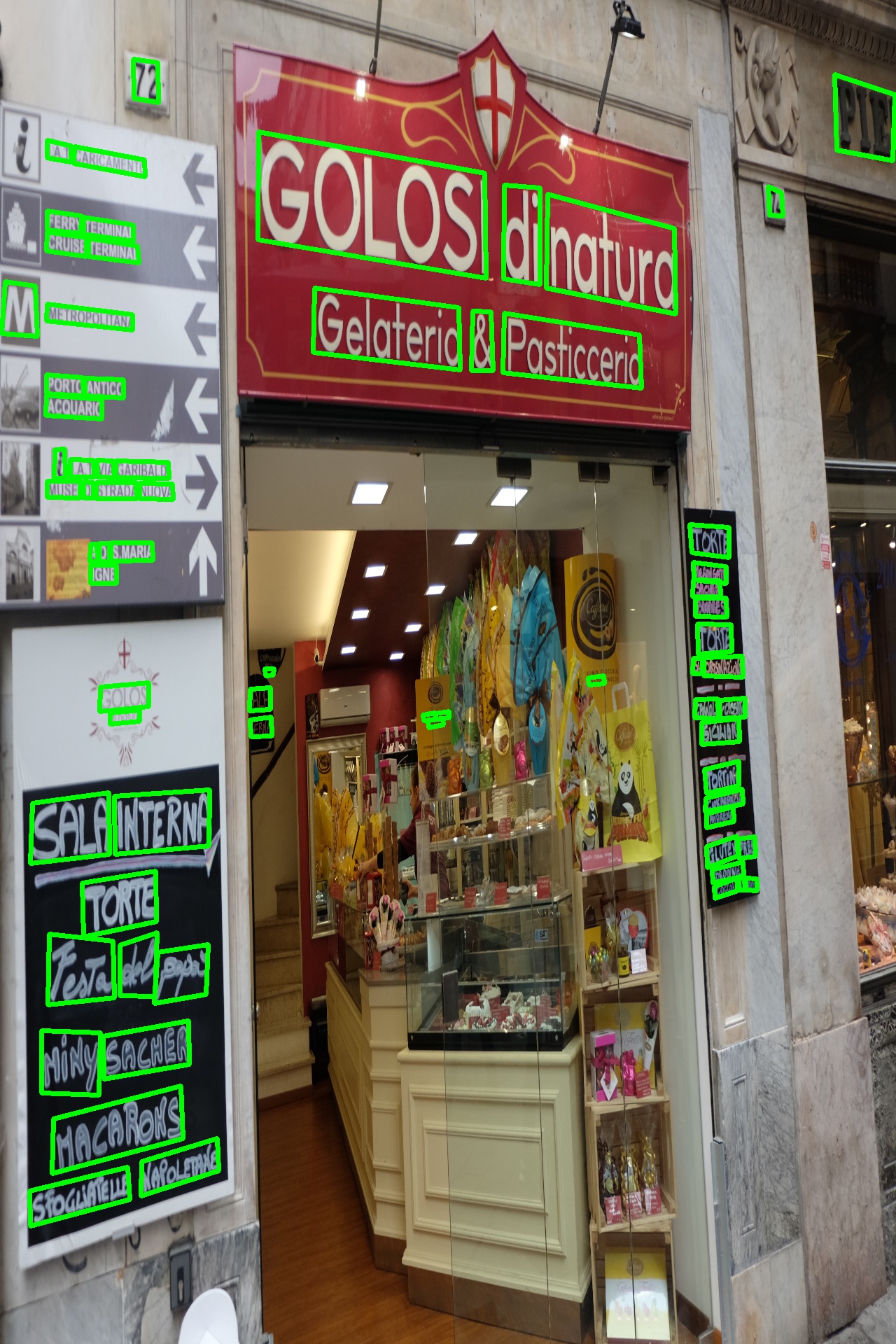}
    \end{minipage}}
    \subfigure[ICDAR 2015]{
    \begin{minipage}[c]{0.3\textwidth}
      \includegraphics[width=\textwidth, height=4cm]{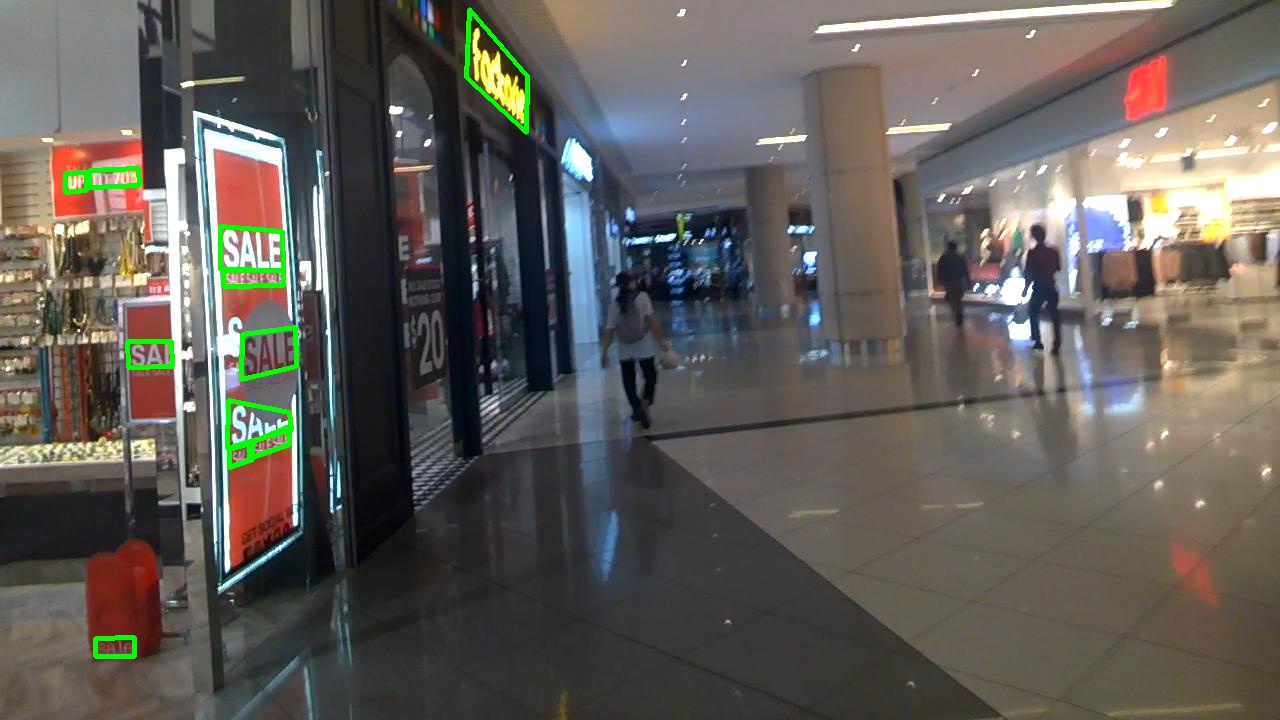}
      \vspace{5pt}
      \includegraphics[width=\textwidth, height=4cm]{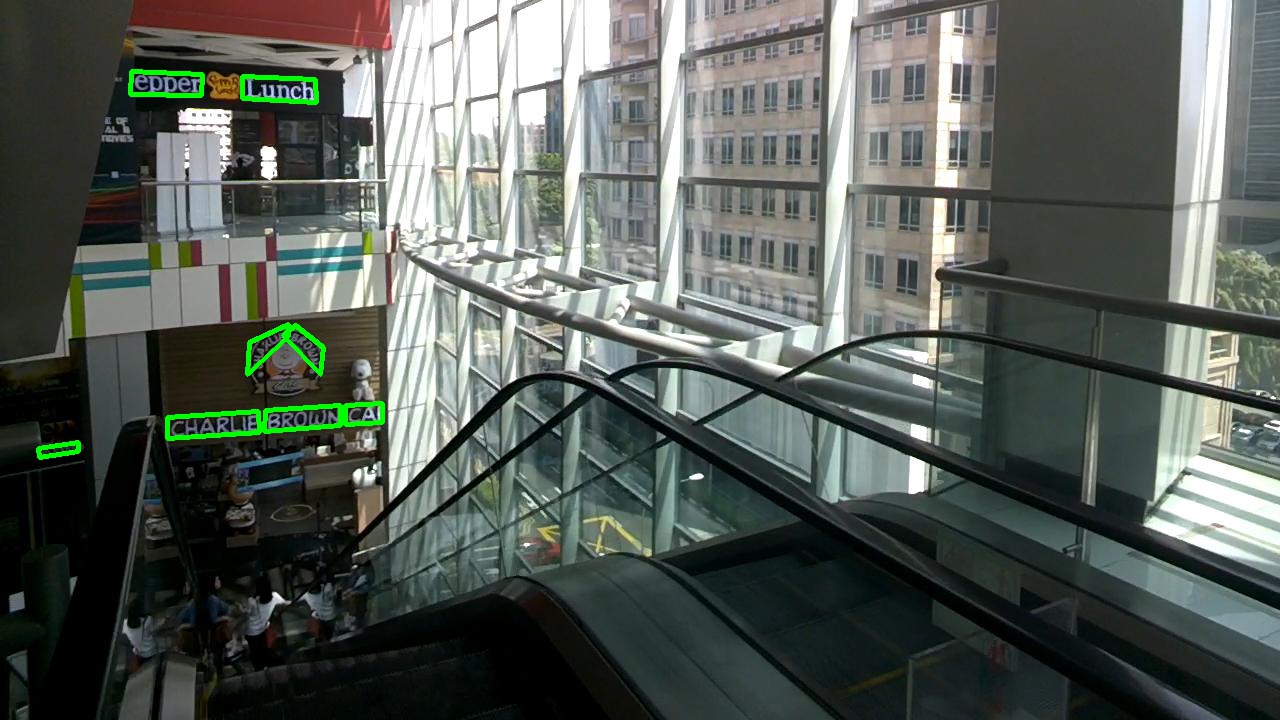}
    \end{minipage}}
    \subfigure[ICDAR 2013]{
    \begin{minipage}[c]{0.3\textwidth}
      \includegraphics[width=\textwidth, height=4cm]{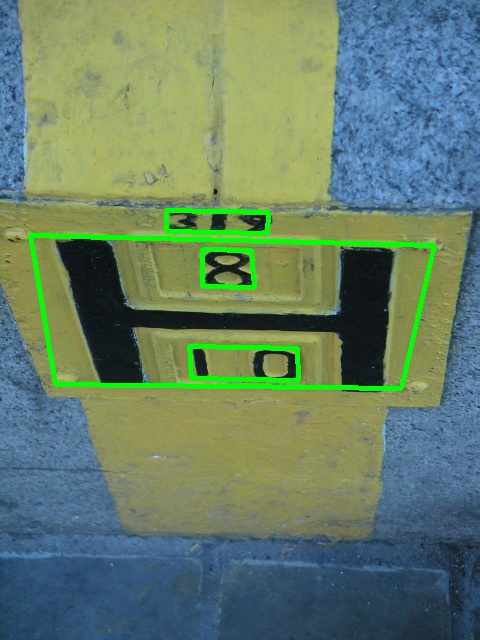}
      \vspace{5pt}
      \includegraphics[width=\textwidth, height=4cm]{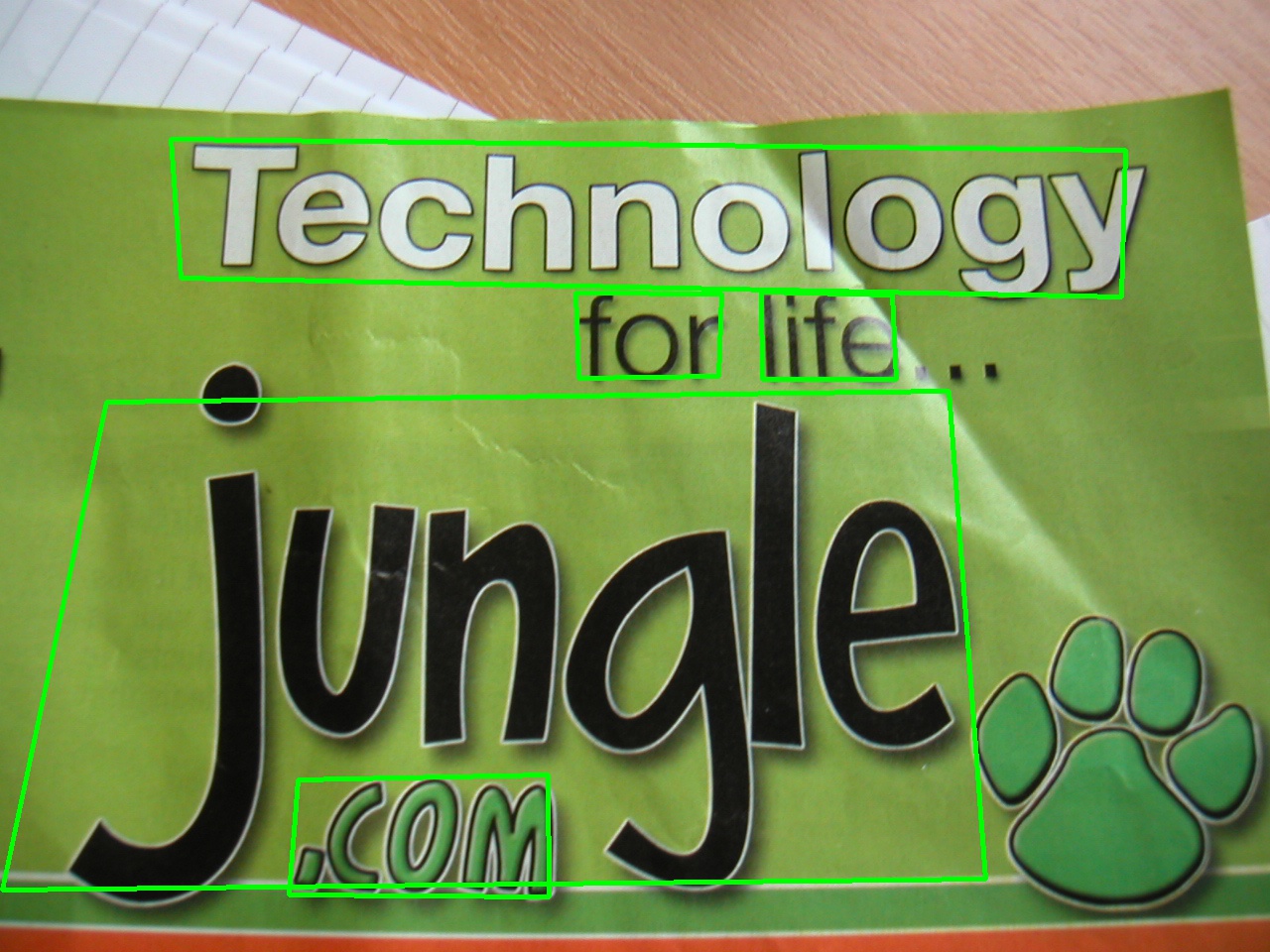}
    \end{minipage}}
    \vspace{5pt}
    \caption{Detection results of PMTD. Best viewed in color.}
    \label{fig:results}
  \end{figure*}

\textbf{ICDAR 2017 MLT}: ImageNet~\cite{deng2009imagenet} pre-trained ResNet50 is adapted to initialize network parameter. We train our model using ICDAR 2017 MLT training and validation images for 160 epochs. We use SGD as our optimizer with batch size 64. The initial learning rate is 0.08 and decays to one-tenth of the previous at the 80th and 128th epoch. During the training stage, images are cropped to $640 \times 640$ patches as described in Sec.~\ref{subsec:our-baseline}. Results are shown in Tab.~\ref{tab:ic17}. For single scale testing, with resizing images' long side to 1600, PMTD achieves an F-measure of 78.48\%. We also resize the long side to 1600 and 2560 for multi-scale testing, and it achieves 80.13\% F-measure, which outperforms the state-of-the-art method by 5.83\%. Qualitative results are shown in Fig.~\ref{fig:results}.

\begin{table}
  \small
  \begin{center}
  \begin{tabular}{l|ccc}
  \hline
  Method    & Precision   & Recall   & F-measure \\ \hline
  SegLink~\cite{shi2017detecting} & 73.10 & 76.80 & 75.00 \\
  SSTD~\cite{he2017single} & 80.00  & 73.00 & 77.00 \\
  WordSup~\cite{hu2017wordsup} & 79.33 & 77.03 & 78.16 \\
  EAST$^*$~\cite{zhou2017east} & 83.27 & 78.33 & 80.72   \\
  R2CNN~\cite{jiang2017r2cnn}  & 85.62 & 79.68 & 82.54 \\
  DDR~\cite{he2017deep} & 82.00 & 80.00 & 81.00 \\
  Lyu \etal$^*$~\cite{lyu2018multi} & 89.50 & 79.70 & 84.30    \\
  RRD$^*$~\cite{liao2018rotation} & 88.00 & 80.00 & 83.80 \\
  TextBoxes++$^*$~\cite{liao2018textboxes++} & 87.80 & 78.50 & 82.90 \\
  PixelLink~\cite{deng2018pixellink} & 85.50 & 82.00 & 83.70 \\
  FOTS~\cite{liu2018fots} & 91.00 & 85.17 & 87.99 \\
  IncepText$^*$~\cite{yang2018inceptext} & 89.40 & 84.30 & 86.80 \\
  TextSnake~\cite{long2018textsnake} & 84.90 & 80.40 & 82.60 \\
  FTSN~\cite{dai2018fused} & 88.60 & 80.00 & 84.10 \\
  SPCNET~\cite{xie2018scene} & 88.70 & 85.80 & 87.20    \\
  PSENet~\cite{li2018shape} & 89.30 & 85.22 & 87.21    \\  \hline
  Baseline & 85.84&	90.55&	88.14 \\
  PMTD & 91.30	&87.43&	\bf{89.33}  \\  \hline
  \end{tabular}
  \end{center}
  \caption{Comparison with other results on ICDAR 2015. $^*$ means multi scale testing. For PMTD, we only report single scale testing result.}
  \label{tab:ic15}
  \end{table}

  \begin{table}
    \small
    \begin{center}
    \begin{tabular}{l|ccc}
    \hline
    Method    & ICDAR13 Eval   & DetEval \\ \hline
    CTPN~\cite{tian2016detecting} & 85.00 & 86.00 \\
    SegLink~\cite{shi2017detecting} & - & 85.30 \\
    TextBoxes$^*$~\cite{liao2017textboxes} & 85.00 & 86.00 \\
    SSTD~\cite{he2017single} & 87.00 & 88.00 \\
    WordSup~\cite{hu2017wordsup} & - & 90.34 \\
    R2CNN~\cite{jiang2017r2cnn}  & 87.73 &- \\
    DDR~\cite{he2017deep} & - & 86.00  \\
    MCN~\cite{liu2018learning} & 88.00 &- \\
    Lyu \etal$^*$~\cite{lyu2018multi} & 88.00 &-  \\
    RRD$^*$~\cite{liao2018rotation} & 89.00 &-\\
    TextBoxes++$^*$~\cite{liao2018textboxes++} &88.00 & 89.00 \\
    PixelLink$^*$~\cite{deng2018pixellink} & - & 88.10 \\
    FEN$^*$~\cite{zhang2018feature}  &91.60 &92.30 \\
    FOTS$^*$~\cite{liu2018fots}  & 92.50 & 92.82 \\
    SPCNET~\cite{xie2018scene} & 92.10 &- \\ \hline
    Baseline &91.73&92.25   \\
    PMTD & \bf{93.40} &\bf{93.59}  \\ \hline
    \end{tabular}
    \end{center}
    \caption{Comparison with other results on ICDAR 2013. $^*$ means multi scale testing. For PMTD, we only report single scale testing result.}
    \label{tab:ic13}
    \end{table}

\textbf{ICDAR 2015}: For ICDAR 2015, we use the pre-trained model from  ICDAR 2017 MLT, and finetune another 40 epochs using ICDAR 2015 training data. Learning rate is set to 0.0008 and unchanged during training. For testing, images' long side are resized to 1920. As shown in Tab.~\ref{tab:ic15}, PMTD outperforms all other methods and achieves 1.19\% higher F-measure than our baseline, which demonstrates the proposed PMTD brings consistent gain on different datasets.

\textbf{ICDAR 2013}: Similar to ICDAR 2015, we finetune 40 epochs on ICDAR 2017 MLT pre-trained model using ICDAR 2013 training data, with fixed 0.0008 learning rate. We resize images' long size to 960 during testing. As shown in Tab.~\ref{tab:ic13}, PMTD surpasses all previous methods once more and gains 1.67\% improvement to the baseline on this dataset.

\begin{table}
  \small
  \begin{center}
  \begin{tabular}{l|ccc}
  \hline
  Method    & Precision   & Recall   & F-measure \\ \hline
  Baseline & 84.72 &	70.37 &	76.88   \\
  Baseline+DC+BU & 85.17 &	70.75 &	77.29    \\ \hline
  PMTD & 85.15&72.77&	\bf{78.48}   \\ \hline
  \end{tabular}
  \end{center}
  \caption{Results of our models with different settings on ICDAR 2017 MLT dataset. PMTD clearly outperforms our baseline.}
  \label{tab:ablation}
\end{table}

\subsection{Ablation Study}\label{subsec:ablation-study}

In this section, we conduct a series of comparative experiments. Experiment results show that our method achieves better performance and predicts more accurate text boxes.

\textbf{Better performance}: We first compare the performance of the baseline and PMTD on ICDAR 2017 MLT dataset. Results are shown in Tab.~\ref{tab:ablation}.

Baseline: Mask R-CNN baseline as described in Sec.~\ref{subsec:our-baseline}, is a solid baseline that significantly outperforms state-of-the-art methods.

Baseline+DC+BU: As described in Sec.~\ref{subsec:pyramid-mask}, we use dilated convolution (DC) and bilinear upsampling (BU) to predict more accurate soft text mask. We also use these two parts for our baseline to measure their gains. Experiment result indicates that dilated convolution and bilinear upsampling only increase baseline by 0.41\%.

PMTD: Our proposed method. It achieves an improvement of 1.6\% compared with the baseline. And in Sec.~\ref{subsec:comparisons-with-other-methods}, experiments on ICDAR 2013 and ICDAR 2015 show PMTD brings consistent gains.

\textbf{More accurate prediction}: As mentioned in Sec.~\ref{sec:introduction}, with the help of the informative soft text mask, the plane clustering algorithm can regress more accurate text boundary and is more robust to imprecise predicted bounding box. However, the current evaluation cannot clearly reflect these two advantages due to the moderate evaluation (only require $IoU \ge 0.5$ on ICDAR 2015 and ICDAR 2017 MLT). So we evaluate PMTD and baseline under a higher IoU threshold.

\begin{figure}
  \includegraphics[width=0.47\textwidth]{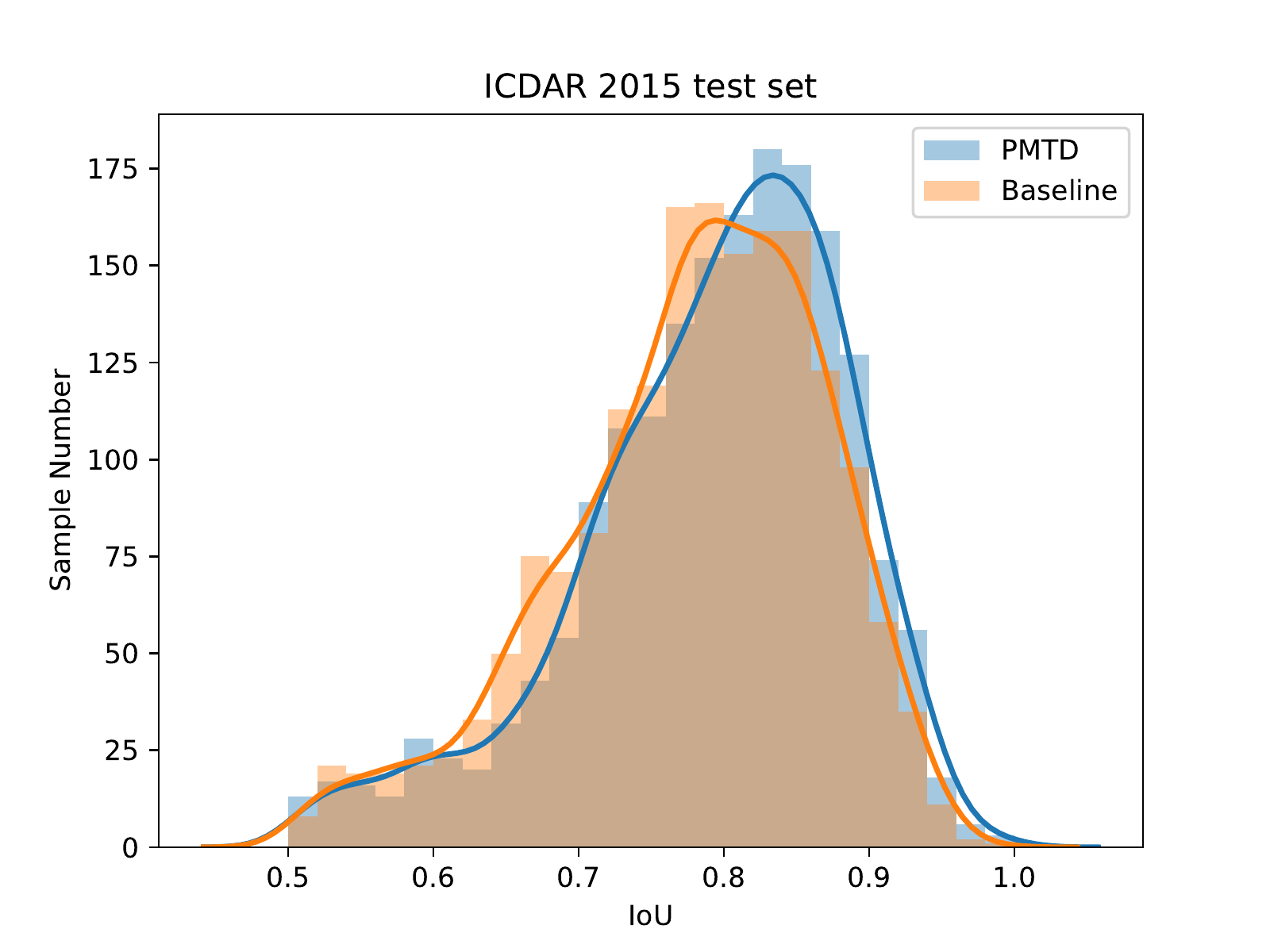}
  \vspace{5pt}
  \caption{Distribution of predicted text boxes with different IoU. PMTD clearly predicts more accurate text boxes than baseline.}
  \label{fig:iou}
\end{figure}

Experiments are constructed on ICDAR 2015 test set for the absence of the label for ICDAR 2017 MLT test set. Results are summarized in Tab.~\ref{tab:iou}. We can see PMTD outperforms baseline by a larger margin when the IoU threshold is 0.8. Especially, PMTD increases F-measure by 18.14\% under 0.8 IoU threshold. Distribution of the true positive samples in different IoU are also illustrated in Fig.~\ref{fig:iou}, which indicates a denser distribution in the high IoU interval for the PMTD.

\begin{table}
\newcommand{\tabincell}[2]{\begin{tabular}{@{}#1@{}}#2\end{tabular}}
  \small
  \begin{center}
  \setlength{\tabcolsep}{1mm}{
  \begin{tabular}{l|ccc|ccc}
  \hline
  \multirow{2}*{IoU}&\multicolumn{3}{c|}{Matched number}&\multicolumn{3}{c}{F-measure} \\ \cline{2-7}
~  &Baseline&PMTD&\tabincell{c}{Relative\\improve}&Baseline&PMTD&\tabincell{c}{Relative\\improve}\\ \hline
  0.5 &   1784 &  1816& 1.79\%&   88.14\%&89.33\%&1.35\% \\
  0.6 &   1696 &  1729& 1.95\%&   83.60\%&84.79\%&1.42\% \\
  0.7 &   1443 & 1556 &7.83\%&   70.44\%&75.31\%&6.91\% \\
  0.8 & 799    &  962  & \bf{20.40}\%&   38.36\%&45.32\%&\bf{18.14}\% \\
  0.9 & 107    &   157  & \bf{46.73}\%&   5.14\%&6.73\%&\bf{30.93}\% \\ \hline
  \end{tabular}}
  \end{center}
  \caption{Number of true positives and F-measure under different IoU threshold on ICDAR 2015. PMTD outperforms baseline significantly when IoU threshold is high.}
  \label{tab:iou}
\end{table}

\begin{figure}
  \includegraphics[width=0.47\textwidth]{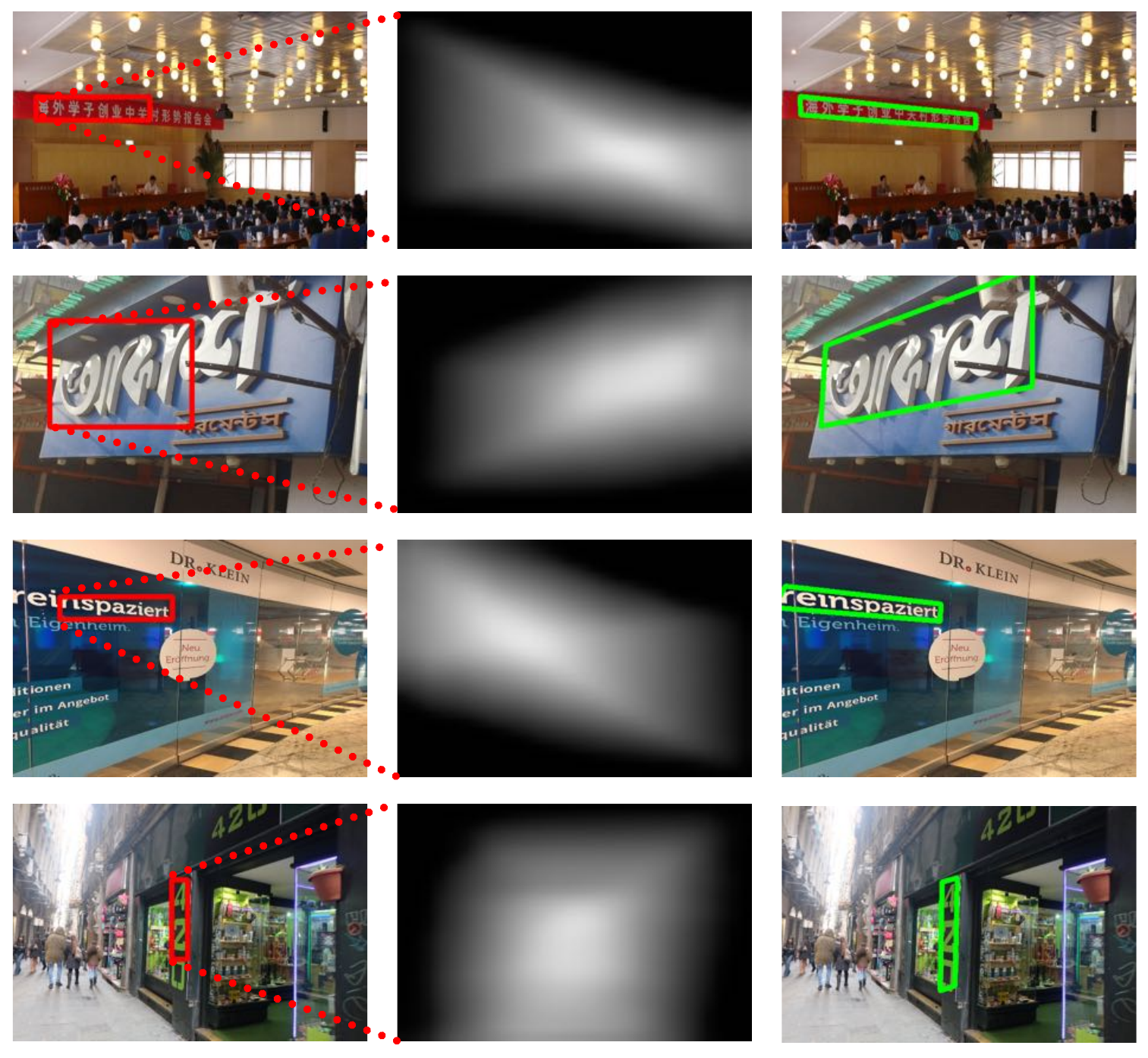}
  \vspace{5pt}
  \caption{PMTD is robust to imprecise predicted bounding box. From left to right: imprecise bounding box, predicted soft text mask, regression text box. It is worth noting that the soft text mask contains gradient information, which helps plane clustering algorithm to regress text box correctly.}
  \label{fig:bounding_box}
\end{figure}

Qualitative results are also shown in Fig.~\ref{fig:results} and Fig.~\ref{fig:bounding_box}. From Fig.~\ref{fig:results} we can see that PMTD can predict satisfactory text boxes, especially for text regions with strange shapes such as trapezoids and curves. Moreover, thanks to the gradient information provided by the soft text mask, PMTD shows the robustness to imprecise predicted bounding boxes as shown in the Fig.~\ref{fig:bounding_box}.

    \section{Conclusion}\label{sec:conclusion}
    In this work, we presented the Pyramid Mask Text Detector (PMTD), which encodes the shape and location information into the supervision and predicts a soft text mask for each text instance. A plane clustering algorithm is introduced to find the most fitting pyramid mask of the predicted soft text mask. Experiments on standard datasets demonstrate the effectiveness of our method.

    {\small
    \bibliographystyle{ieee}
    \bibliography{egbib}
    }

\end{document}